\documentclass{article}
\usepackage{geometry}
\usepackage{microtype}
\usepackage{float}
\usepackage{hyperref}
\usepackage{amsmath,amssymb,amsthm}
\usepackage{commath}
\usepackage{graphicx}
\usepackage{color}
\usepackage{array}
\usepackage[linesnumbered,boxed,vlined]{algorithm2e}
\usepackage{enumerate}
\usepackage{enumitem}
\usepackage{hyperref}
\usepackage{url}
\usepackage{caption}
\usepackage{cleveref}

\usepackage[export]{adjustbox}
\usepackage{multirow}
\usepackage{tensor}
\usepackage[caption=false]{subfig}
\usepackage{empheq}
\usepackage{booktabs} 
\usepackage[table,xcdraw]{xcolor}
\usepackage{bm}

\newcolumntype{a}{>{\columncolor{color-1}}c}
\newcolumntype{b}{>{\columncolor{color-2}}c}

\newcommand{\HumanDynamicspath}{./Res/Images/HumanDynamics}

\begin{document}

\title{A Graph Neural Network Approach for Temporal Mesh Blending and Correspondence}

\author{\normalsize Aalok Gangopadhyay \quad Abhinav Narayan Harish \quad Prajwal Singh \quad Shanmuganathan Raman\\
\normalsize \textit{ CVIG Lab, Indian Institute of Technology Gandhinagar}\\
\normalsize  \{aalok, abhinav.narayan, singh\_prajwal,  shanmuga\}@iitgn.ac.in}
\date{}


\maketitle

\section{Introduction}
\label{HMD:sec:introduction}

When we observe a human performing a motion, such as walking, running, crawling, or jumping, it is easy for us to recognize and distinguish them. By observing an action sequence, even if it is short-duration, our internal understanding of motion enables us to estimate the action sequence in the near past or near future. This understanding helps us in our daily life, for instance, while moving in a crowded place. However, imparting the ability to understand motion to computers is challenging due to the complex dynamics involved. It is a problem of great significance in various fields, such as human-computer interaction, sports analytics, rehabilitation, and surveillance. 

Over the past few decades, this problem has garnered much attention. Due to advances in motion capture technologies, it has become possible to capture and store human motion digitally. In recent years, computer vision techniques for understanding human dynamics, such as depth-based motion analysis \cite{ye2013survey}, 3D skeleton-based action classification \cite{presti20163d}, optical and scene flow-based warping method \cite{li2019dense, zablotskaia2019dwnet}, and attention-based human motion transfer \cite{zhu2019progressive}, have shown substantial progress. Several works, such as \cite{kanazawa2019learning, kocabas2020vibe}, have developed methods for extracting pose and motion from videos and have achieved impressive results. However, the progress in processing the temporal sequence of shapes has been limited due to the lack of temporal shape sequence datasets. 

In this work, we address the shape blending problem, where given the shape of a person at two different time instants, the task is to estimate the mesh at an arbitrary time instant in the past, in the future, or between the two instants. The motivation behind solving this problem is that if an algorithm can temporally interpolate or extrapolate the motion given a few frames as input, it has an implicit understanding of human motion. The information present in this algorithm can then be used for further downstream tasks. Temporal blending for human skeletons have been addressed recently in \cite{harvey2020robust, holden2016deep, holden2015learning, barsoum2018hp, chiu2019action, fragkiadaki2015recurrent, gopalakrishnan2019neural, jain2016structural, martinez2017human, pavllo2020modeling}. In contrast, this work focuses on the temporal blending of human meshes instead of skeletons. Moreover, we do not assume the meshes to be in correspondence, which makes the problem even more challenging. 

In this work, we have proposed a self-supervised deep learning framework for solving the shape blending problem. The following are the major contributions of this work:
\begin{enumerate}[noitemsep,nolistsep]
\item We have utilized existing motion capture datasets and human mesh-generating algorithms to create a large dataset of temporal mesh sequences of humans performing various real-world motions.
\item We have proposed a graph neural network along with a conditional refinement scheme for solving the mesh correspondence problem for aligning the input meshes.
\item We have proposed a graph neural network for solving the mesh blending problem given an arbitrary time instant.
\end{enumerate}

We formally define the problem statement in section \ref{HMD:sec:problem_statement}. The proposed architecture and the dataset generation process are explained in section \ref{HMD:sec:ProposedApproach} and \ref{HMD:sec:DatasetGeneration}, respectively. Finally, we demonstrate the performance of our approach through qualitative results in section \ref{HMD:sec:results_and_discussion}.

\section{Problem Statement}
\label{HMD:sec:problem_statement}

Let the triangular mesh of a 2-manifold embedded in 3D be represented as $\mathcal{G}=(\mathcal{V},\mathcal{F},\mathcal{A})$. Here, $\mathcal{V}=(v_1,v_2,\cdots,v_n)^\top \in \mathbb{R}^{n\times3}$ denotes the vertex features, $n$ being the number of vertices, $\mathcal{F}$ denotes the set containing $f$ faces, and $\mathcal{A} \in \{0,1\}^{n \times n}$ denotes the adjacency matrix. Let us consider two structurally isomorphic meshes $\mathcal{G}_0=(\mathcal{V}_0,\mathcal{F}_0,\mathcal{A}_0)$ and $\mathcal{G}_1=(\mathcal{V}_1,\mathcal{F}_1,\mathcal{A}_1)$ whose adjacency matrices are related through a permutation matrix $\mathcal{P}$ such that $\mathcal{A}_1 = \mathcal{P}\mathcal{A}_0\mathcal{P}^\top$. Let us further set the convention that $\mathcal{G}_0$ occurs at $t{=}0$ and $\mathcal{G}_1$ occurs at $t{=}1$, where, $t \in \mathbb{R}$ denotes time. Given $\mathcal{G}_0$ and $\mathcal{G}_1$, our goal is to predict $\mathcal{G}_t=(\mathcal{V}_t,\mathcal{F}_t,\mathcal{A}_t)$, for an arbitrary time $t$, such that $\mathcal{G}_t$ is isomorphic to the input meshes $\mathcal{G}_0$ and $\mathcal{G}_1$. The problem of predicting $\mathcal{G}_t$ for a given value of $t$ is referred to as the shape blending problem. Based on the value of $t$, shape blending can further be divided as follows: 
(i) Shape Interpolation $(0 < t < 1)$,
(ii) Future Shape Extrapolation $(t > 1)$, and
(iii) Past Shape Extrapolation $(t < 0)$.
Our primary motivation is understanding human dynamics, so we focus only on blending human meshes. We assume that the input human meshes are triangular, 2-manifold, and watertight.

\begin{figure}[!h]
\begin{center}
  \includegraphics[width=\linewidth]{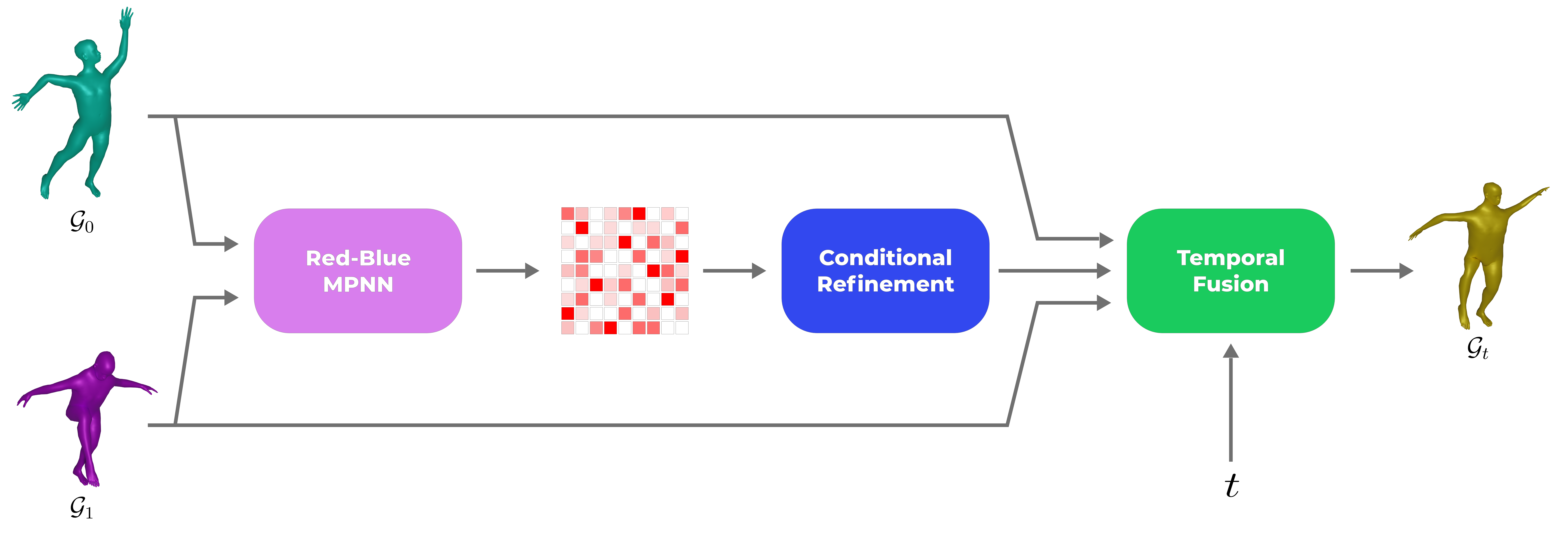}
  \caption{The proposed approach for shape blending. The \textit{Red-Blue MPNN} component takes two isomorphic meshes as input and estimates a soft correspondence. The \textit{Conditional Refinement} component converts the soft correspondence into a hard correspondence which is used to align the two meshes. The \textit{Temporal Fusion} component performs temporal blending on the aligned meshes for an arbitrary value of time $t$.}
  \label{HMD:fig:HMD_Net}
\end{center}
\end{figure}

\section{Proposed Approach}
\label{HMD:sec:ProposedApproach}

Our proposed approach, as depicted in Fig. \ref{HMD:fig:HMD_Net}, consists of three components: (a) \textit{Red-Blue MPNN}, that outputs a soft correspondence matrix estimating the permutation matrix $\mathcal{P}$ between $\mathcal{G}_0$ and $\mathcal{G}_1$, (b) \textit{Conditional Refinement}, where the soft correspondence is converted into a hard correspondence, and under special conditions, a refinement procedure is initiated that obtains the permutation matrix accurately, and (c) \textit{Temporal Fusion}, which utilizes this hard correspondence matrix to perform shape blending and obtains $\mathcal{G}_t$, given an arbitrary time $t$. We describe each of these components in detail.

\subsection{Red-Blue MPNN}
\label{HMD:subsec:Red-Blue MPNN}

\begin{figure}[!h]
\begin{center}
  \includegraphics[width=\linewidth]{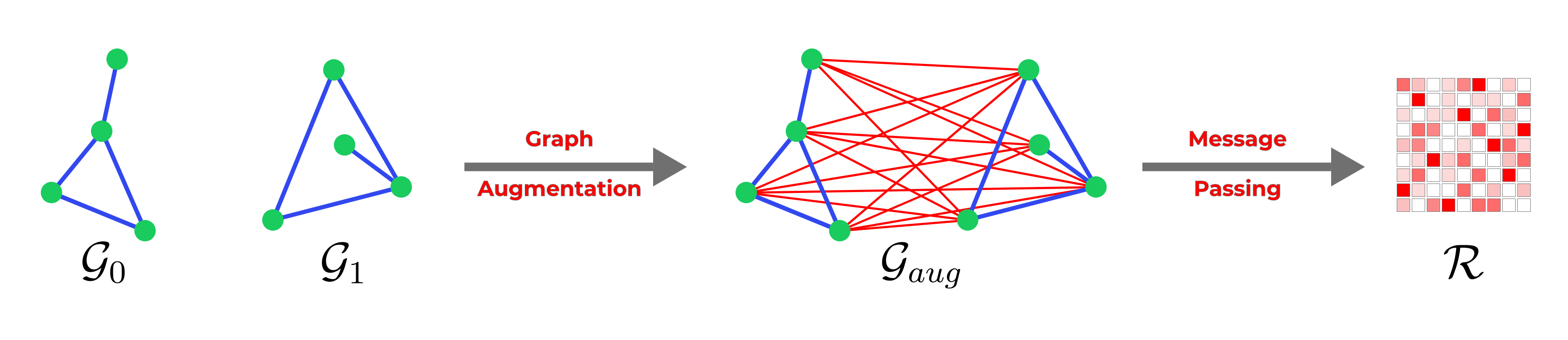}
  \caption{Given two meshes $\mathcal{G}_0$ and $\mathcal{G}_1$, during graph augmentation each vertex of $\mathcal{G}_0$ is connected with each vertex of $\mathcal{G}_1$ through a red colored edge. Message passing is performed on this augmented graph where both the vertices and edges have features. The features of the red edges are normalized and iterpreted as the soft correspondence matrix.}
  \label{HMD:fig:RBMPNN}
\end{center}
\end{figure}

Given $\mathcal{G}_0=(\mathcal{V}_0,\mathcal{F}_0,\mathcal{A}_0)$ and $\mathcal{G}_1=(\mathcal{V}_1,\mathcal{F}_1,\mathcal{A}_1)$,
we construct an augmented graph $\mathcal{G}_{aug}$ formed by two types of weighted edges : \textit{blue-edges} - denoting the original edges in $\mathcal{G}_0$ and $\mathcal{G}_1$ and \textit{red-edges} - linking every vertex of $\mathcal{G}_0$ to that of $\mathcal{G}_1$. The binary-valued blue-edges preserve the structural information present within each of $\mathcal{G}_0$ and $\mathcal{G}_1$. The real-valued red-edges enable the information flow between $\mathcal{G}_0$ and $\mathcal{G}_1$. The adjacency matrix of $\mathcal{G}_{aug}$ is given by $\mathcal{A}_{aug} = \bigg[\begin{smallmatrix}
  \mathcal{A}_0 & \mathcal{R}\\
  \mathcal{R}^\top & \mathcal{A}_1
\end{smallmatrix}\bigg]$, where, $\mathcal{R}$ is an $n \times n$ matrix representing the weights of red-edges. The graph augmentation process is visually illustrated in Fig. \ref{HMD:fig:RBMPNN}.

Message passing is performed on $\mathcal{G}_{aug}$ for $K$ iterations, where the vertex features and red-edges features are updated in each iteration. 
The transpose of the weight matrix of the red-edges obtained at the end of the message passing scheme denoted as ($\mathcal{R}^K)^\top$, is then interpreted as a soft correspondence matrix estimating the permutation matrix $\mathcal{P}$. The weight of the red-edge $R_{ij}$ indicates the probability that the $i^{th}$ vertex in $\mathcal{G}_0$ is a match of the $j^{th}$ vertex in $\mathcal{G}_1$. We design a message passing scheme for processing $\mathcal{G}_{aug}$, which is described in Algorithm \ref{HMD:algo:RBMPNN}.

\begin{algorithm}[!h]
\caption{Red-Blue MPNN}
\label{HMD:algo:RBMPNN}
\SetAlgoLined
\DontPrintSemicolon
 \textbf{Initialization:}\;
  $\mathcal{R}^{(0)}=\frac{1}{n} \cdot \mathcal{\mathbf{1}}_{n \times n}$ \;
  $\mathcal{V}_0^{(0)} = f_{init}(\mathcal{V}_0)$\;
  $\mathcal{V}_1^{(0)} = f_{init}(\mathcal{V}_1)$\;
 \For{$0 \leq i < K$}{
  $\mathcal{V}_0^{(i+1)} = \alpha(\mathcal{A}_0 \mathcal{V}_0^{(i)}W_{\mathcal{B}}^{(i)} + \mathcal{R}^{(i)} \mathcal{V}_1^{(i)}W_{\mathcal{R}}^{(i)})$\;
  $\mathcal{V}_1^{(i+1)} = \alpha(\mathcal{A}_1 \mathcal{V}_1^{(i)}W_{\mathcal{B}}^{(i)} + (\mathcal{R}^{(i)})^{\top} \mathcal{V}_0^{(i)}W_{\mathcal{R}}^{(i)})$\;
  $\hat{\mathcal{R}}^{(i+1)} = (\mathcal{N}(\mathcal{V}_0^{(i+1)})W_{\mathcal{T}})(\mathcal{N}(\mathcal{V}_1^{(i+1)})W_{\mathcal{T}})^{\top}$\;
  $\mathcal{R}^{(i+1)} = \lambda_{\mathcal{S}} \cdot \mathcal{S}(\hat{\mathcal{R}}^{(i+1)}) + \lambda_{\mathcal{R}} \cdot \mathcal{R}^{(i)}$\;
  }
 \textbf{return} $\sigma((\mathcal{R}^K)^\top)$
\end{algorithm}

In Algorithm \ref{HMD:algo:RBMPNN}, $\mathcal{\mathbf{1}}_{n \times n}$ is the $n \times n$ matrix filled with all ones, $n$ being the number of vertices in $\mathcal{G}_0$ and $\mathcal{G}_1$, 
$\mathcal{R}^{(i)}$ indicates the weights of the red-edges updated during the $i^{th}$ iteration, 
$\alpha$ indicates the ReLU activation function and $\sigma$ indicates the sigmoid activation function, 
$f_{init}(x) = \alpha(\alpha(\alpha(xW_1+b_1)W_2+b_2)W_3+b_3)$ is a 3-layer fully connected neural network, 
$W_{\mathcal{B}}^{(i)}$ and $W_{\mathcal{R}}^{(i)}$ indicate the MPNN weights of the blue-edges and the red-edges, respectively, for the $i^{th}$ iteration, 
$\mathcal{V}_0^{(i)}$ and $\mathcal{V}_1^{(i)}$ indicate the vertex features of $\mathcal{G}_0$ and $\mathcal{G}_1$, respectively, updated during the $i^{th}$ iteration, 
$\mathcal{N}(\cdot)$ indicates the $\ell_2$ normalization operator, 
$W_{\mathcal{T}}$ is the weight of a fully connected layer,
$\mathcal{S}(\cdot)$ is the Sinkhorn-normalization operator used for making the estimation close to a doubly stochastic matrix, 
$\lambda_{\mathcal{S}}$ and $\lambda_{\mathcal{R}}$ are the weights used for computing the running average of the estimation.
After $K$ iterations of message passing, the sigmoid activation is applied to the transpose of the red-edges weight matrix, which is the estimate of the permutation matrix $\mathcal{P}$. 



\begin{figure}[]
\begin{center}
  \includegraphics[width=0.8\linewidth]{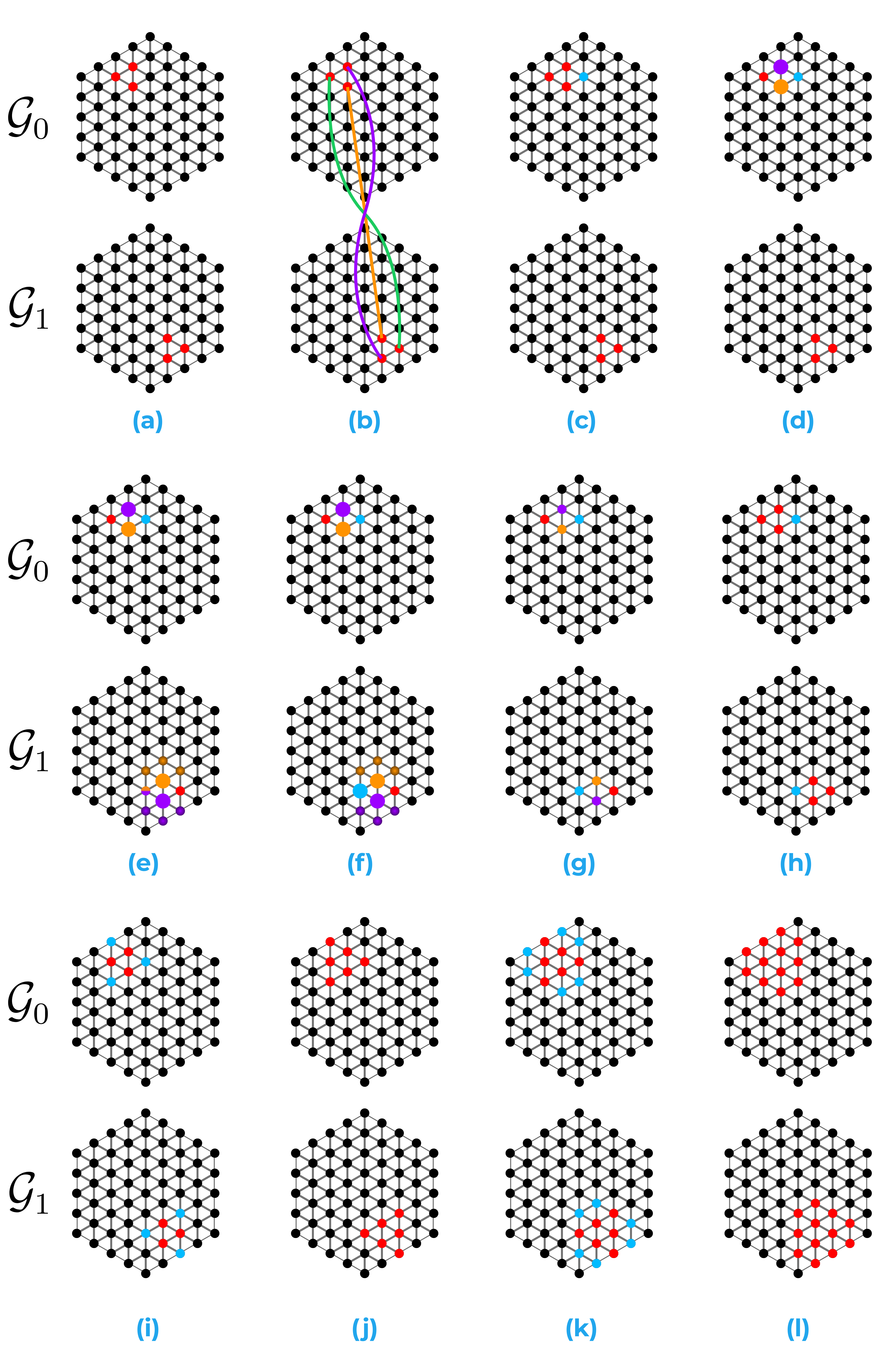}
  \caption{An example illustrating Conditional Refinement. Consider a scenario where (a) triangles $T_0$ and $T_1$ (red) are found in $\mathcal{G}_0$ and $\mathcal{G}_1$, respectively, with (b) their vertex-wise matches. Given (c) a vertex $v^0$ (blue) in $\mathcal{G}_0$, in the neighbourhood of $T_0$, there are two triangle vertices $v_a^0$ (violet) and $v_b^0$ (orange) that are adjacent to $v^0$. (e) The match of $v_a^0$ and $v_b^0$ in the other mesh $\mathcal{G}_1$ are $v_a^1$ and $v_b^1$. The neighbors of $v_a^1$ and $v_b^1$ are marked in their respective colors. (f) The match of $v^0$ in the other mesh $\mathcal{G}_1$ would lie in the shared neighbourhood of $v_a^1$ and $v_b^1$. There is exactly one vertex in the shared neighbourhood due to the meshes being isomorphic. Thus, (g) we find $v_1$ (blue), the match of $v_0$ in $\mathcal{G}_1$. Thus, (h) starting from $T_0$, we refined the correspondence by finding the match of a neighboring vertex. (i) This process applied to all the vertices in the 1-hop neighbourhood of $T_0$, (j) helps us obtain the neighbourhood matches. (k) Applying this process iteratively enables us (l) to expand the triangle seed into bigger patches.}
  \label{HMD:fig:IP_illustration}
\end{center}
\end{figure}

\begin{figure}[]
\begin{center}
  \includegraphics[width=0.8\linewidth]{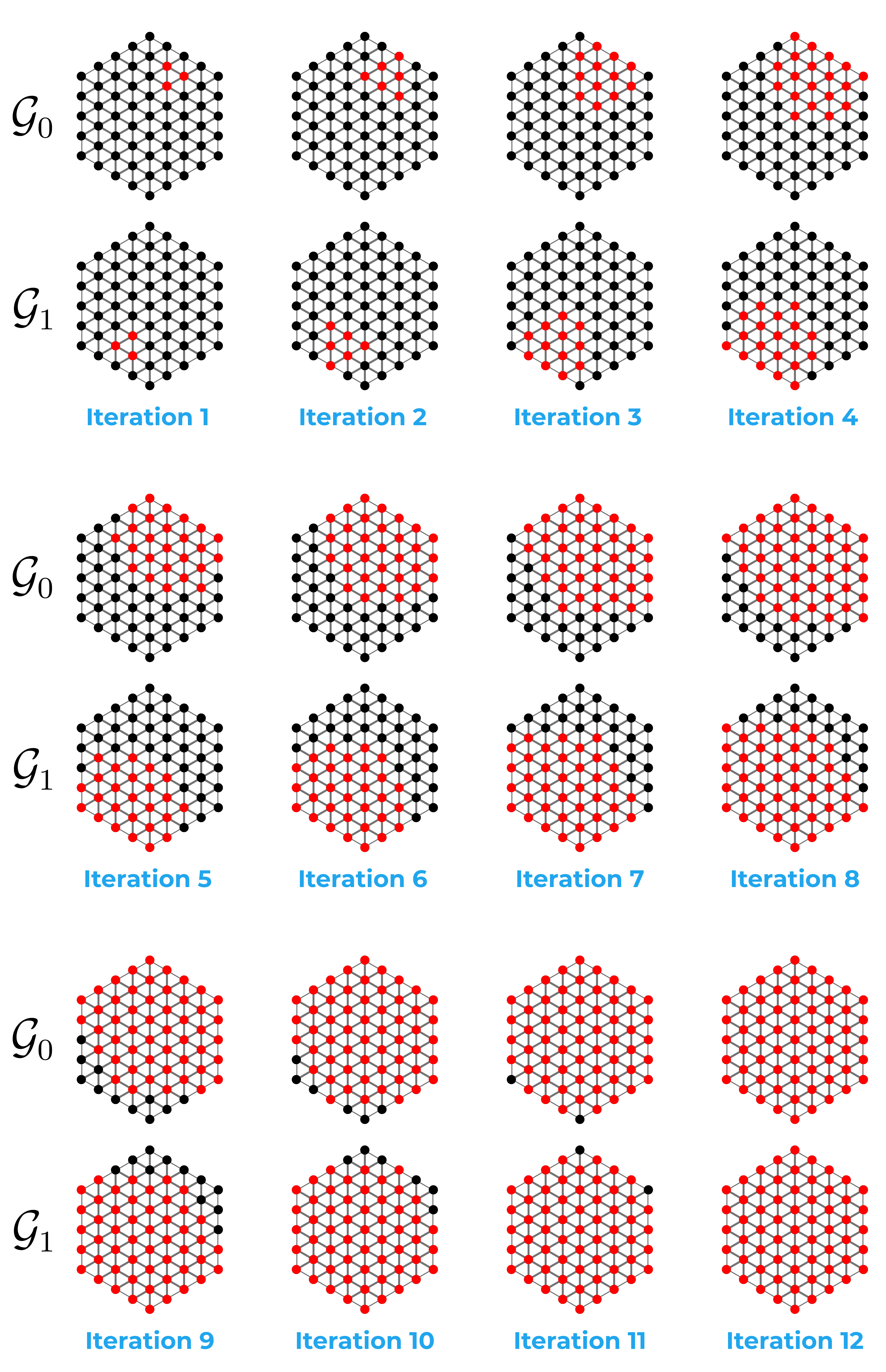}
  \caption{Once a triangle is found in both $\mathcal{G}_0$ and $\mathcal{G}_1$ that is in correspondence, the refinement process is applied iteratively to find the correspondences in the entire mesh.}
  \label{HMD:fig:IP_Iterations}
\end{center}
\end{figure}

\subsection{Conditional Refinement}
\label{HMD:subsec:Conditional Refinement}

The output of the \textit{Red-Blue MPNN} is a soft correspondence matrix with real values between $0$ and $1$. In order to use it for temporal shape blending, we need to convert it to a binary matrix that indicates the hard correspondences. To obtain this, we perform binarization, converting the column entry with the largest value to one for each row and the rest of the column entries to zero. Although this makes the matrix binary, it does not guarantee that the matrix is a valid permutation matrix. Considering that the input shapes $\mathcal{G}_0$ and $\mathcal{G}_1$ are 2-manifold triangular meshes, there is a particular condition, that, when satisfied, the exact permutation matrix can be recovered. If such a condition is satisfied, the \textit{Conditional Refinement} process is initiated, and the exact permutation matrix obtained is the final output. Otherwise, the binary matrix obtained after binarization is the output. We now describe the condition under which the refinement process is initiated.

Let us define $\mathcal{H}_{01} = \textit{bin}(\mathcal{R}^K)$ and $\mathcal{H}_{10} = \textit{bin}((\mathcal{R}^K)^\top)$, where $\textit{bin}$ is the binarization operator described above.
If binarization results in the accurate estimation of the permutation matrix upto automorphisms, then we would have, 
\begin{equation}
\label{HMD:eq:matching}
\mathcal{A}_1 = \mathcal{H}_{10}\mathcal{A}_0\mathcal{H}_{10}^\top \quad \text{and} \quad
\mathcal{A}_0 = \mathcal{H}_{01}\mathcal{A}_1\mathcal{H}_{01}^\top
\end{equation}
In cases where the estimation is not accurate, equality would not hold for the entire adjacency matrix as in \ref{HMD:eq:matching}. However, there might be rows or columns in the adjacency matrix for which the equality holds, which allows us to find partial vertex matches between $\mathcal{G}_0$ and $\mathcal{G}_1$. If we find a triangle in the mesh, such that all the three vertices in that triangle are correctly matched, then this allows us to propagate the matches further along the mesh as depicted in Fig. \ref{HMD:fig:IP_illustration} and Fig. \ref{HMD:fig:IP_Iterations}. 

Let us say we find such a triangle in both graphs. Lets denote them as $T_0 = (v_a^0,v_b^0,v_c^0)$ in $\mathcal{G}_0$ and its match $T_1 = (v_a^1,v_b^1,v_c^1)$ in $\mathcal{G}_1$. Consider a vertex $v^0$ in the neighbourhood of $T_0$, whose match in $\mathcal{G}_1$ is $v^1$. Since $\mathcal{G}_0$ and $\mathcal{G}_1$ are isomorphic, $v^1$ would be in the neighbourhood of $T_1$.
Since $\mathcal{G}_0$ is triangular, watertight, and 2-manifold, at least two of the vertices in $T_0$ would be adjacent to $v^0$. Let us say, without loss of generality, $v_a^0$ and $v_b^0$ are the adjacent vertices. Then we know that the match of $v^0$ lies in the shared neighbourhood of the matches of $v_a^0$ and $v_b^0$. That is, $v^1$ is in the neighbour of both $v_a^1$ and $v_b^1$. However, there cannot be more than one vertex in the shared neighbourhood of $v_a^1$ and $v_b^1$, which allows us to find $v^1$, the match of $v^0$.

\begin{figure}[!h]
\begin{center}
  \includegraphics[width=\linewidth]{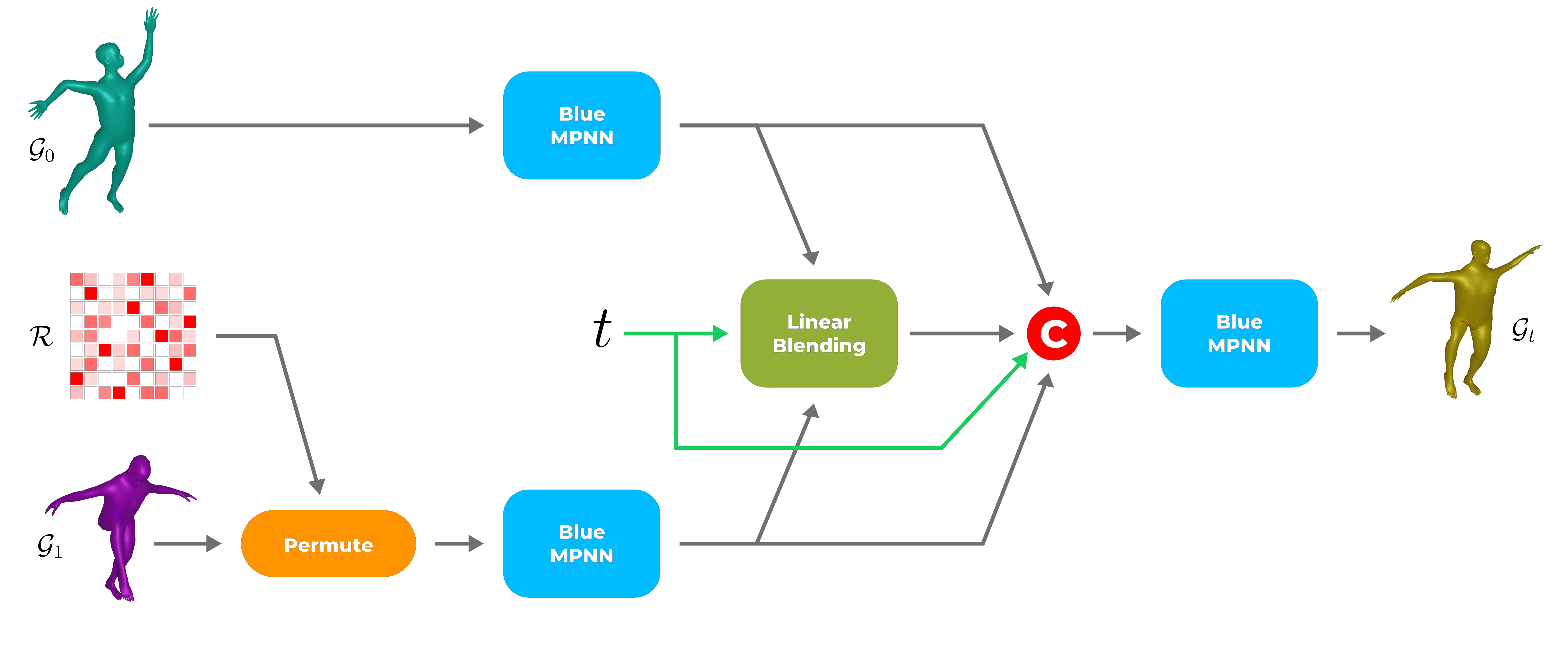}
  \caption{After the hard correspondence is estimated, the two meshes are brought into alignment followed by a process of temporal fusion which obtains the temporal blending between the two input meshes.}
  \label{HMD:fig:TemporalFusion}
\end{center}
\end{figure}

\subsection{Temporal Fusion}
\label{HMD:subsec:Temporal Fusion}

In temporal fusion, we first align the two meshes $\mathcal{G}_0$ and $\mathcal{G}_1$, which are permuted versions of each other.
Suppose the exact permutation matrix $\mathcal{P}$ is recovered during \textit{Conditional Refinement}. In that case, we permute all the vertices of $\mathcal{G}_1$ to align it with the vertices of $\mathcal{G}_0$. The reordered vertices of $\mathcal{G}_1$ can be written as $\mathcal{P}^{\top}\mathcal{V}_1$.
Otherwise, we reorder only those matched vertices if only partial matches are found. We estimate the coordinates for the remaining vertices using a propagation process, where we propagate the coordinates from the matched vertices to the neighouring unmatched vertices. The coordinate of an unmatched vertex in the neighbourhood of matched vertices is updated as the mean of the coordinates of all matched vertices in the neighbourhood. This way, the alignment is obtained, although some shape information might get lost. 
Once the alignment is found, we proceed with the fusion process, which involves the \textit{Blue MPNN} (\textit{BMPNN}), which performs message passing in the graph and updates the vertex features. \textit{BMPNN} takes as input the vertex features $\mathcal{V}$ and the adjacency matrix $\mathcal{A}$ and outputs the updated vertex features given by  
$\textit{BMPNN}(\mathcal{V},\mathcal{A}) =   h(h(h(h(h(h(((\alpha(\mathcal{V}W_1 + b_1))W_2 + b2),\mathcal{A}),\mathcal{A}),\mathcal{A}),\mathcal{A}),\mathcal{A}),\mathcal{A})$, where 
$h(\mathcal{V},\mathcal{A}) = g(\mathcal{V},\mathcal{A}) + \mathcal{V}$ in which 
$g(\mathcal{V},\mathcal{A}) = \alpha(\hat{\mathcal{D}}^{-\frac{1}{2}}\hat{\mathcal{A}}\hat{\mathcal{D}}^{-\frac{1}{2}}\mathcal{V}W)$. 
Here, $\alpha$ is the ReLU activation function, $W$, $W_1$, $W_2$ are learnable weight matrices, $b_1$ and $b_2$ are biases, $\hat{\mathcal{A}} = \mathcal{A} + \mathcal{I}$ ($\mathcal{I}$ being the identity matrix), and $\hat{\mathcal{D}}$ is the diagonal vertex degree matrix of $\hat{\mathcal{A}}$.

We first use \textit{BMPNN} to compute the vertex features $\hat{\mathcal{V}}_{0}$ and $\hat{\mathcal{V}}_{1}$ of both the meshes. Given time $t$, we then compute the linear blending $\hat{\mathcal{V}}_{t}^{LB} = (1-t)\hat{\mathcal{V}}_{0} + (t)\hat{\mathcal{V}}_{1}$. We then obtain $\hat{\mathcal{V}}_{t} = (\hat{\mathcal{V}}_{0} || \hat{\mathcal{V}}_{1} || \hat{\mathcal{V}}_{t}^{LB} || t \cdot \mathcal{\mathbf{1}}_{n \times d})$ through concatenation, where $d$ is the length of the feature vector. We finally perform message passing on these concatenated features to obtain $\mathcal{G}_t=(\mathcal{V}_t,\mathcal{A}_t)$, 
where $\mathcal{V}_t = \textit{BMPNN}(\hat{\mathcal{V}}_{t},\mathcal{A}_0)$  and $\mathcal{A}_t = \mathcal{A}_0$.

\section{Dataset Generation and Training}
\label{HMD:sec:DatasetGeneration}

We first use MakeHuman \cite{makehuman} to create 1000 human character meshes of different shapes by randomly varying the parameters that control the shape. 
We utilize the CMU motion capture dataset \cite{cmumocap}, which contains a total of 2605 instances of human motions spanning 23 categories. 
Each human mesh generated in MakeHuman is equipped with a skeletal armature compatible with the skeleton in the CMU motion capture dataset. 
For each character and motion capture sequence, we use the retargeting algorithm available in blender \cite{blender} to create a motion sequence of meshes. We generate a dataset containing $2605000$ temporal sequences of meshes, each sequence having a different number of frames. We divide the dataset into a training set (80\%) and a testing set (20\%).
The advantage of using MakeHuman is that all characters have the same mesh connectivity. The adjacency graphs are isomorphic, and only the vertex locations change when the shape is deformed.

\subsection{Mesh Correspondence}

For training \textit{Red-Blue MPNN}, during each iteration, we randomly choose a mesh motion sequence and sample a frame from it at random to obtain the mesh $\mathcal{G}_0$. We choose a random permutation $\mathcal{P}$ and apply it to $\mathcal{G}_0$ to obtain $\mathcal{G}_1$. We forward pass $\mathcal{G}_0$ and $\mathcal{G}_1$ through \textit{Red-Blue MPNN} to obtain the estimation $\hat{\mathcal{P}}$. We utilize the loss function mentioned in equation \ref{HMD:eq:corrloss} to train \textit{Red-Blue MPNN}.
\begin{equation}
\label{HMD:eq:corrloss}
\mathcal{L} = || \hat{\mathcal{P}} \mathcal{A}_0 \hat{\mathcal{P}}^{\top} - \mathcal{A}_1 ||_2 + || \hat{\mathcal{P}}^{\top}  \mathcal{A}_1 \hat{\mathcal{P}}- \mathcal{A}_0 ||_2  + || \hat{\mathcal{P}}\hat{\mathcal{P}}^{\top} - \mathcal{I} ||_2 + || \hat{\mathcal{P}}^{\top}\hat{\mathcal{P}} - \mathcal{I} ||_2
\end{equation}

\subsection{Mesh Blending}

During each training iteration, we choose a mesh motion sequence at random. From that sequence we sample three different frames to obtain $\mathcal{G}_a$, $\mathcal{G}_b$, and $\mathcal{G}_c$, their frame indices being $t_a$, $t_b$, and $t_c$, respectively, with $t_a < t_b < t_c$. In each iteration, we train our network for the following three tasks: 
\begin{enumerate}[noitemsep,nolistsep]
\item \textit{mesh interpolation}: $\mathcal{G}_a$, $\mathcal{G}_c$, and $t=\frac{t_b - t_a}{t_c - t_a}$ as input, $\mathcal{G}_b$ as output
\item \textit{future mesh extrapolation}: $\mathcal{G}_a$, $\mathcal{G}_b$, and $t=\frac{t_c - t_a}{t_b - t_a}$ as input, $\mathcal{G}_c$ as output
\item \textit{past mesh extrapolation}: $\mathcal{G}_b$, $\mathcal{G}_c$, and $t=\frac{t_a - t_b}{t_c - t_b}$ as input, $\mathcal{G}_a$ as output
\end{enumerate}
We use a random permutation for each task to reorder the second input mesh.
We use the loss function mentioned in equation \ref{HMD:eq:chamfer_loss} to train \textit{Temporal Fusion} for all three tasks.
\begin{multline}
  \mathcal{L}_{chamfer} = \frac{1}{|\mathcal{V}_{pred}|}\sum_{u \in \mathcal{V}_{pred}}\min_{v \in \mathcal{V}_{gt}}\norm{\mathcal{P}_{pred}(u) - \mathcal{P}_{gt}(v)}_2^2  \\ + \frac{1}{|\mathcal{V}_{gt}|}\sum_{v \in \mathcal{V}_{gt}}\min_{u \in \mathcal{V}_{pred}}\norm{\mathcal{P}_{pred}(u) - \mathcal{P}_{gt}(v)}_2^2 
  \label{HMD:eq:chamfer_loss}
\end{multline}
Here $\mathcal{V}_{pred}$ and $\mathcal{V}_{gt}$ are the predicted and ground-truth mesh vertices, respectively.



\section{Results}
\label{HMD:sec:results_and_discussion}

\begin{figure}[!p]
\begin{center}
  \includegraphics[width=\linewidth]{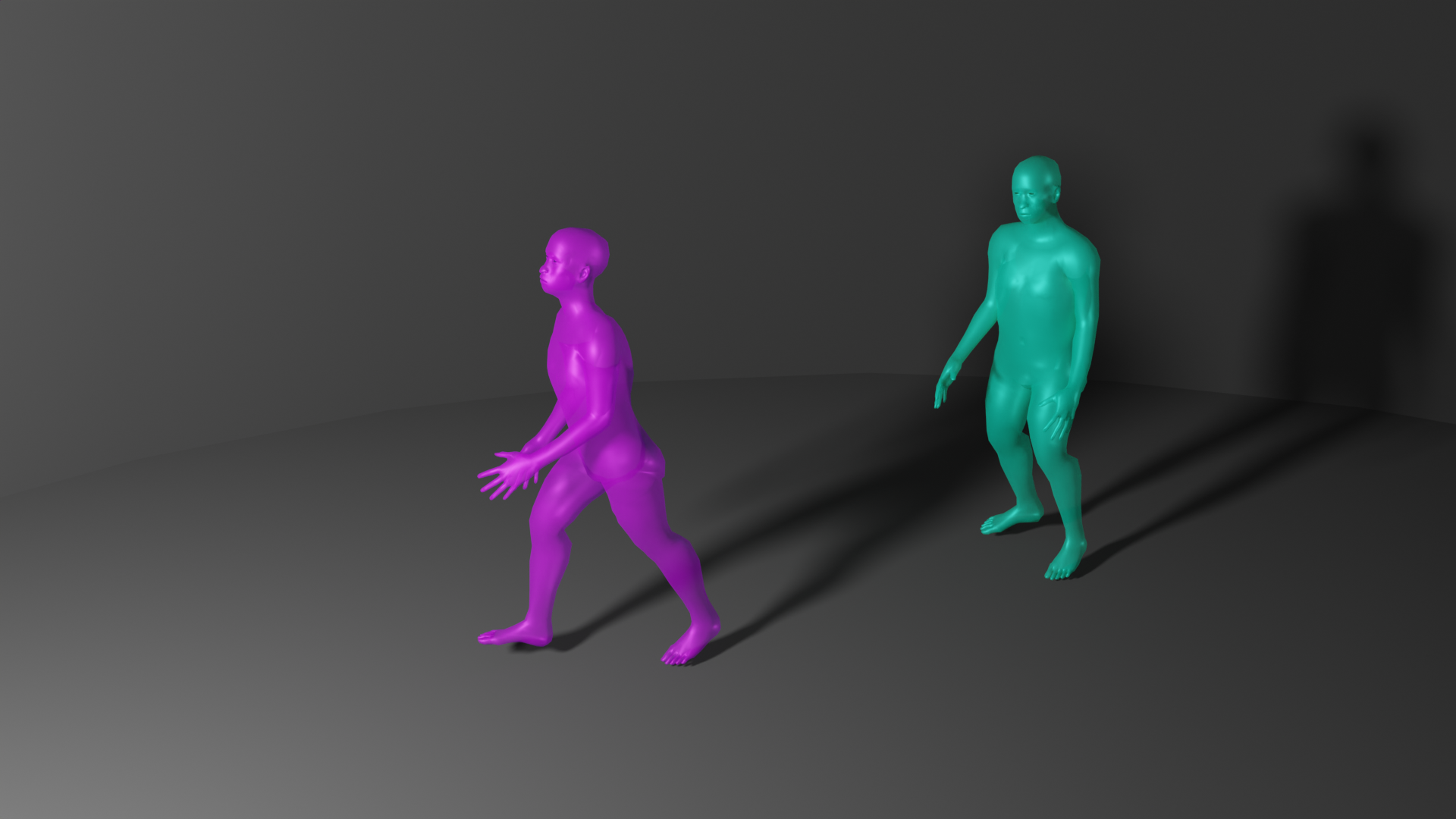}
  \includegraphics[width=\linewidth]{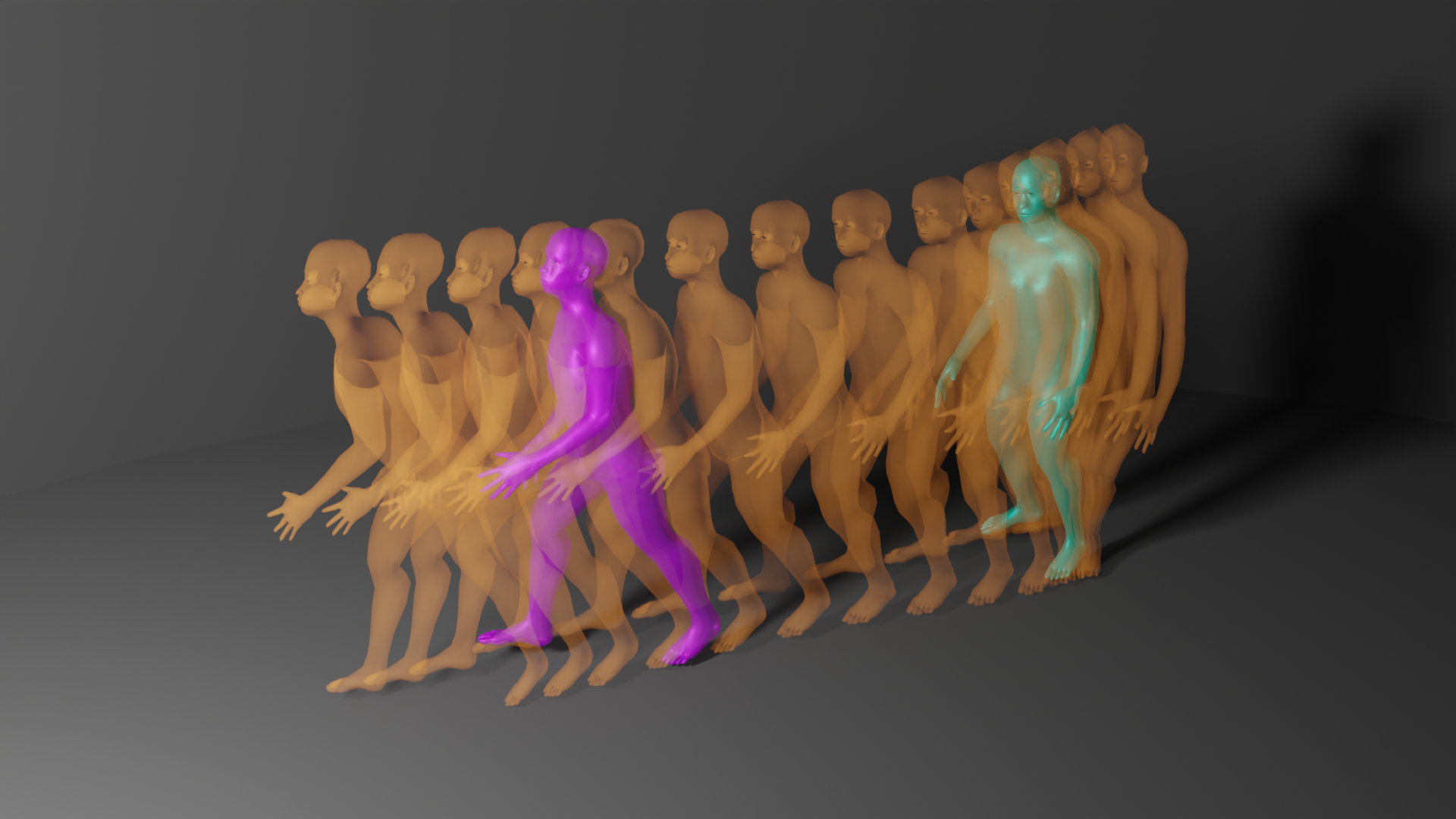}
  \caption{Qualitative Results: A sequence of a person bending forward and taking a right turn while walking.}
  \label{HMD:fig:Results_1}
\end{center}
\end{figure}

\begin{figure}[!p]
\begin{center}
  \includegraphics[width=\linewidth]{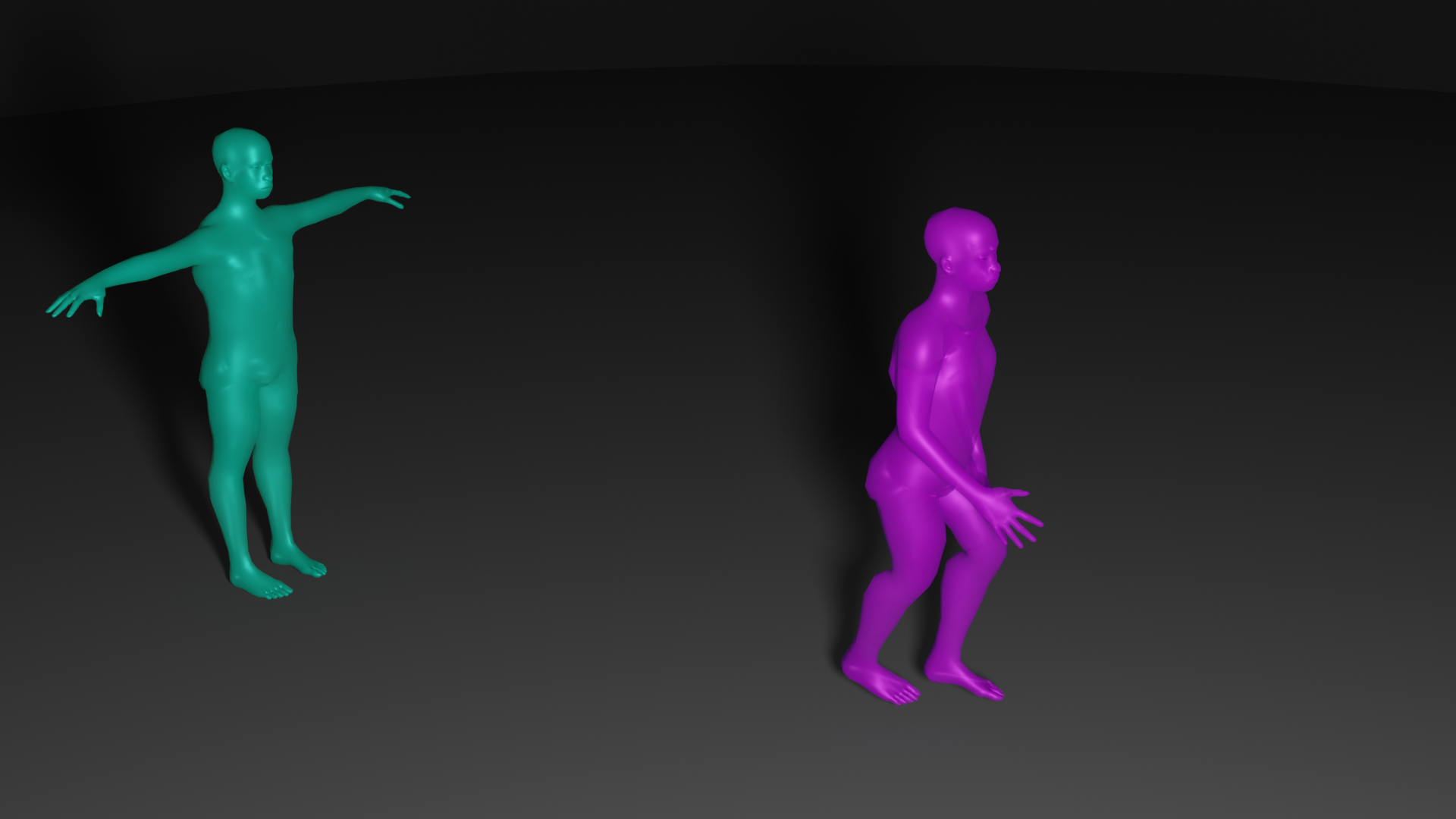}
  \includegraphics[width=\linewidth]{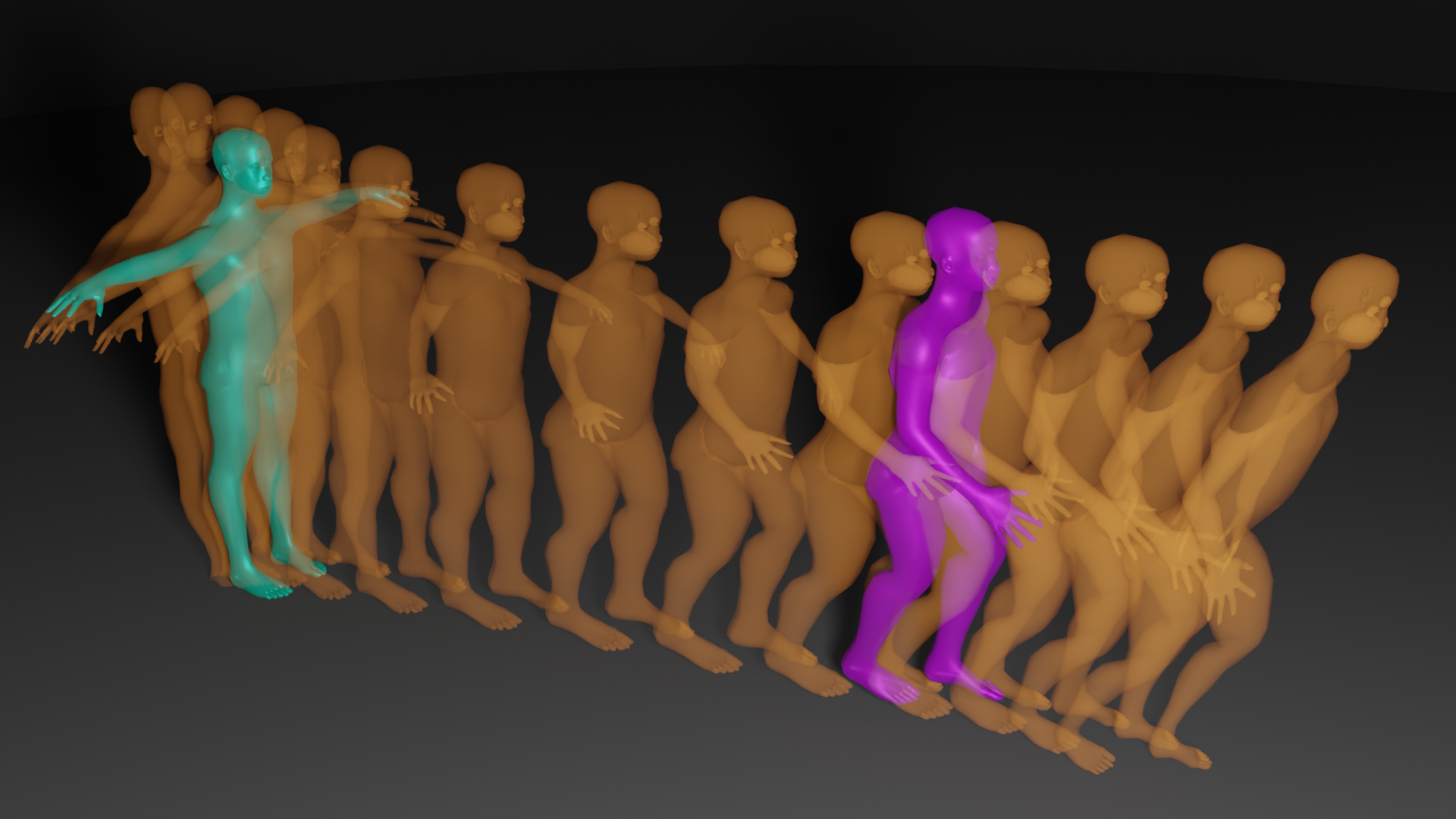}
  \caption{Qualitative Results: A sequence of a person with arms wide open walks and moves their arms inwards.}
  \label{HMD:fig:Results_2}
\end{center}
\end{figure}

\begin{figure}[!p]
\begin{center}
  \includegraphics[width=0.48\linewidth]{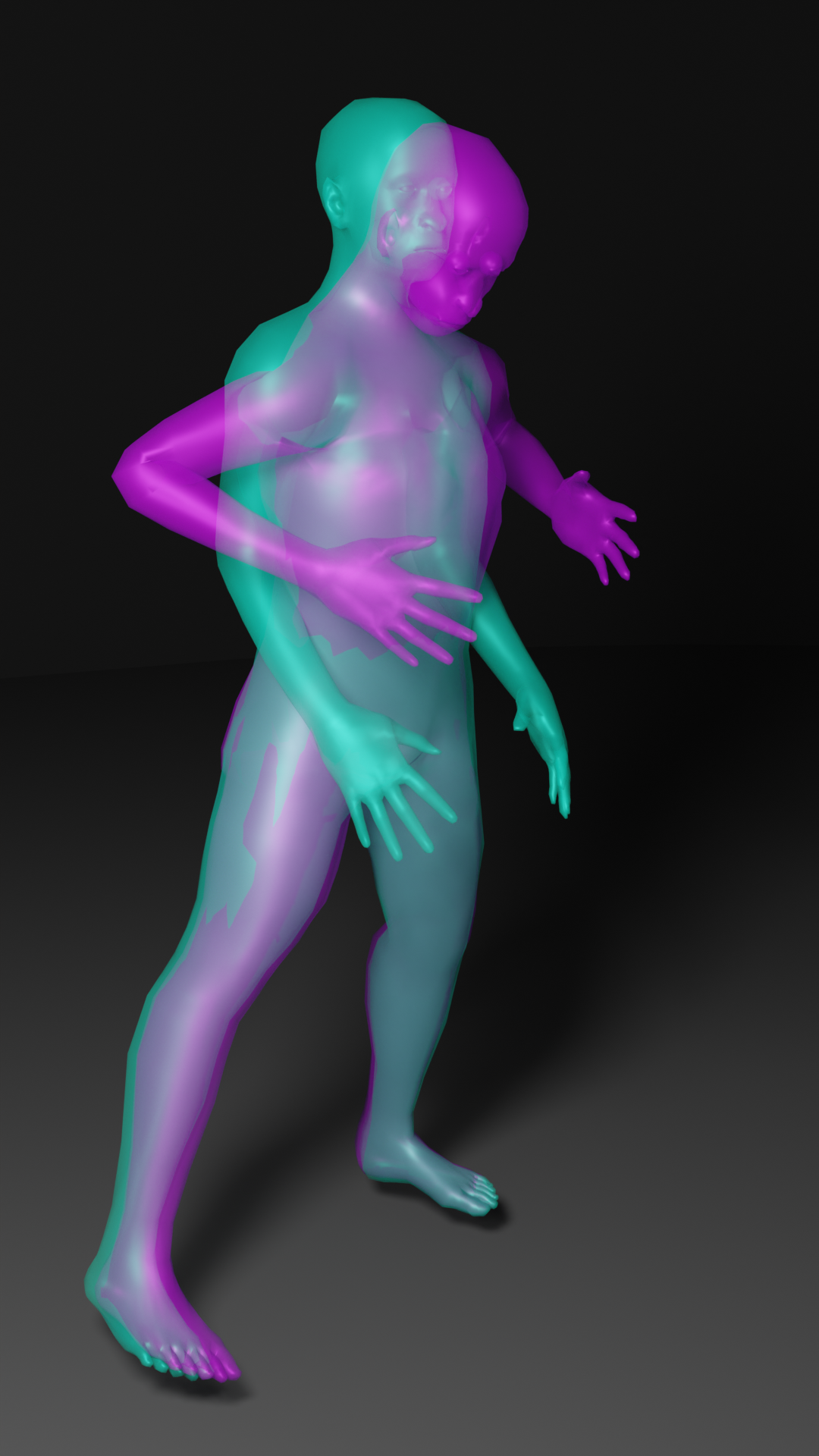}
  \includegraphics[width=0.48\linewidth]{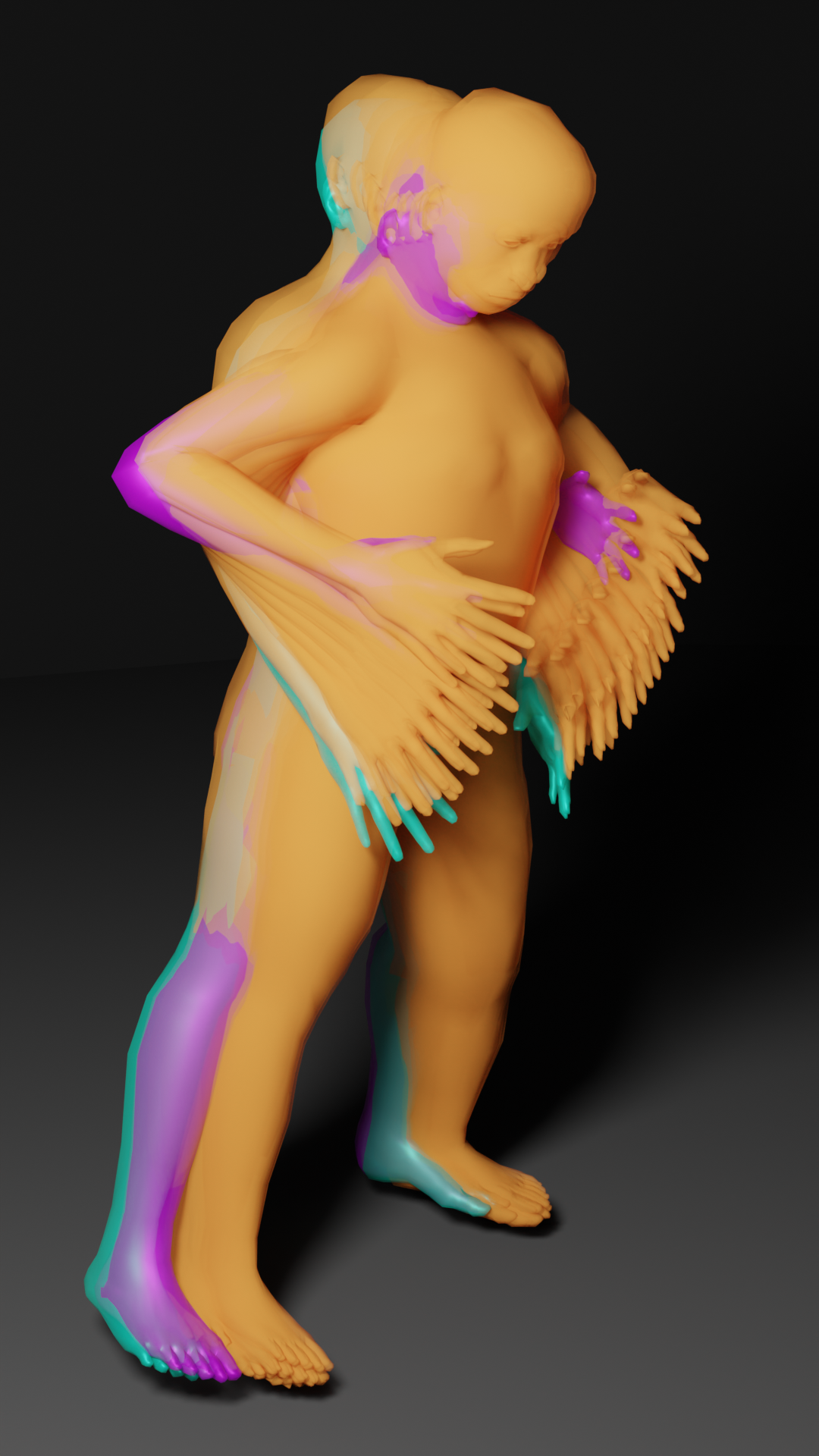}
  \caption{Qualitative Results: A sequence of a person standing in one place and making complicated hand movements.}
  \label{HMD:fig:Results_3}
\end{center}
\end{figure}

\begin{figure}[!p]
\begin{center}
  \includegraphics[width=\linewidth]{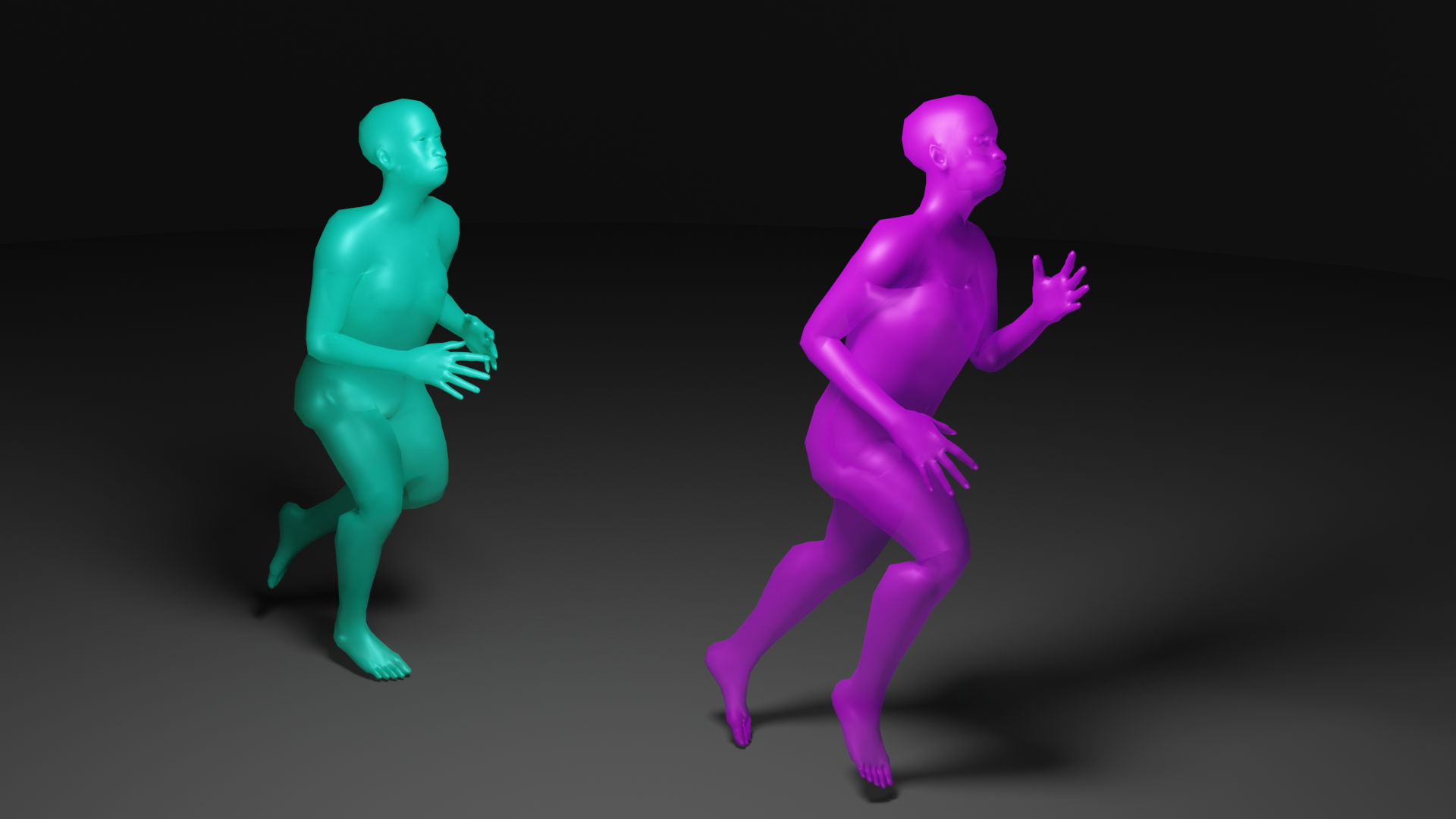}
  \includegraphics[width=\linewidth]{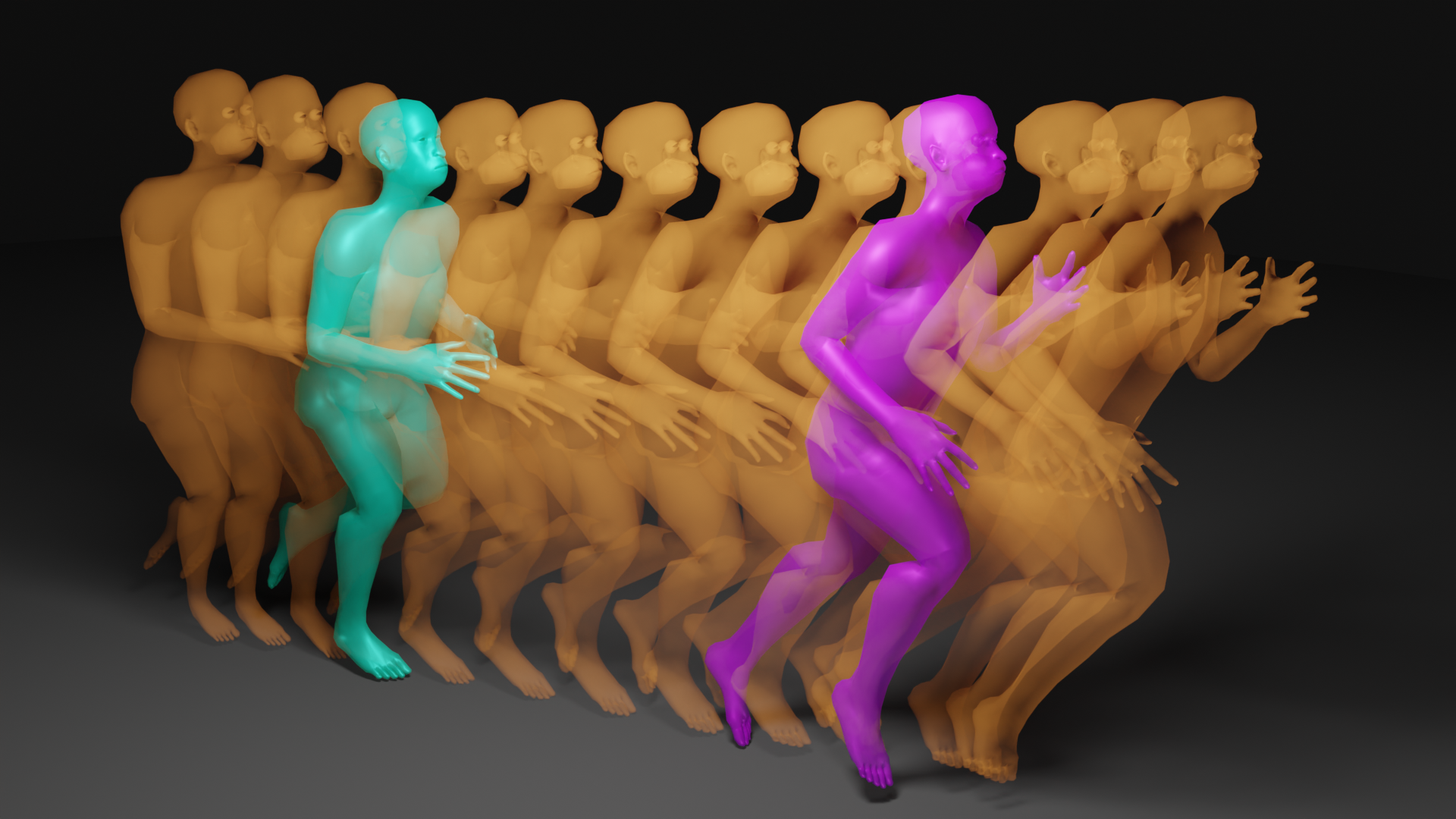}
  \caption{Qualitative Results: A sequence of a person running.}
  \label{HMD:fig:Results_4}
\end{center}
\end{figure}

\begin{figure}[!p]
\begin{center}
  \includegraphics[width=0.48\linewidth]{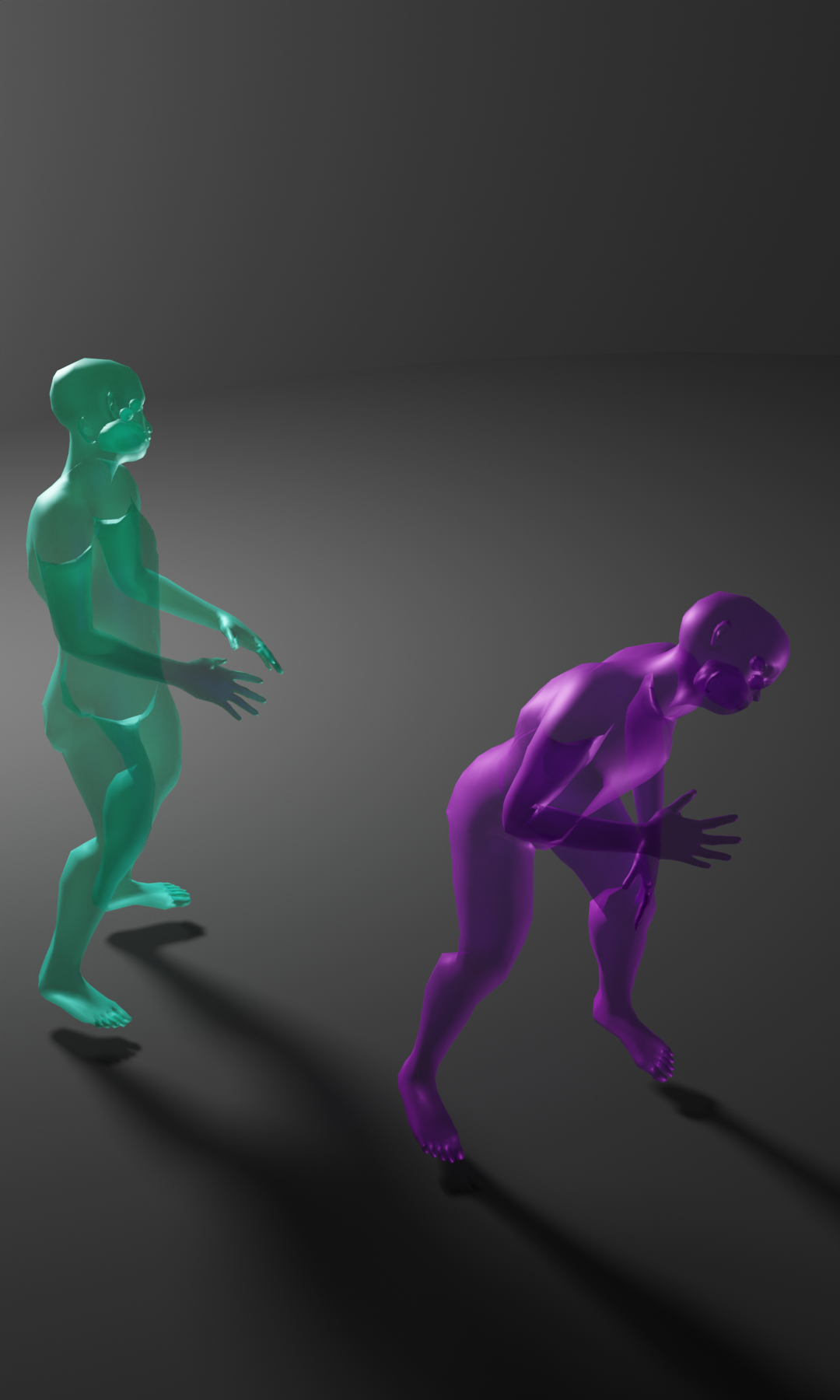}
  \includegraphics[width=0.48\linewidth]{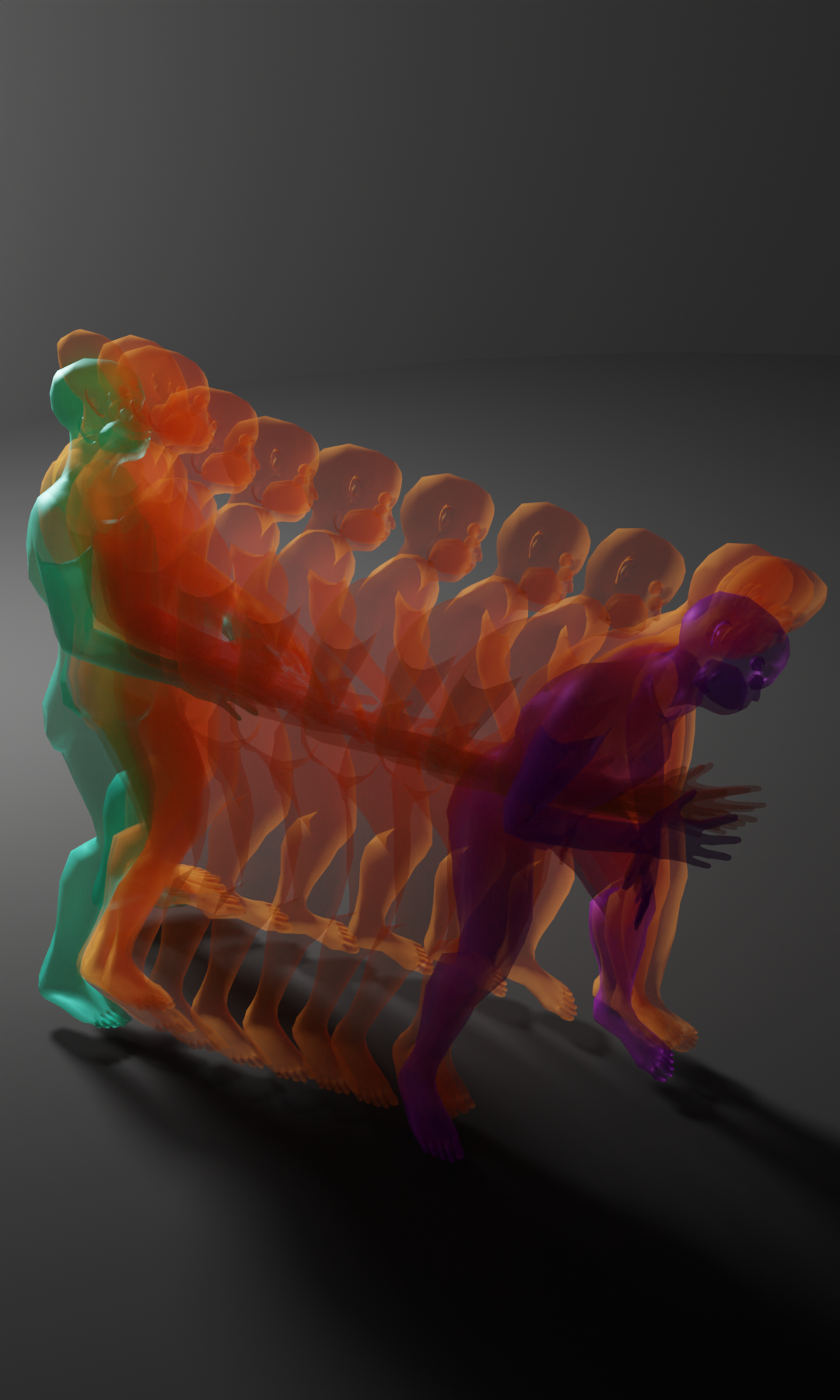}
  \caption{Qualitative Results: A sequence of a person bending forward with large hand and leg movements.}
  \label{HMD:fig:Results_5}
\end{center}
\end{figure}

\begin{figure}[!p]
\begin{center}
  \includegraphics[width=\linewidth]{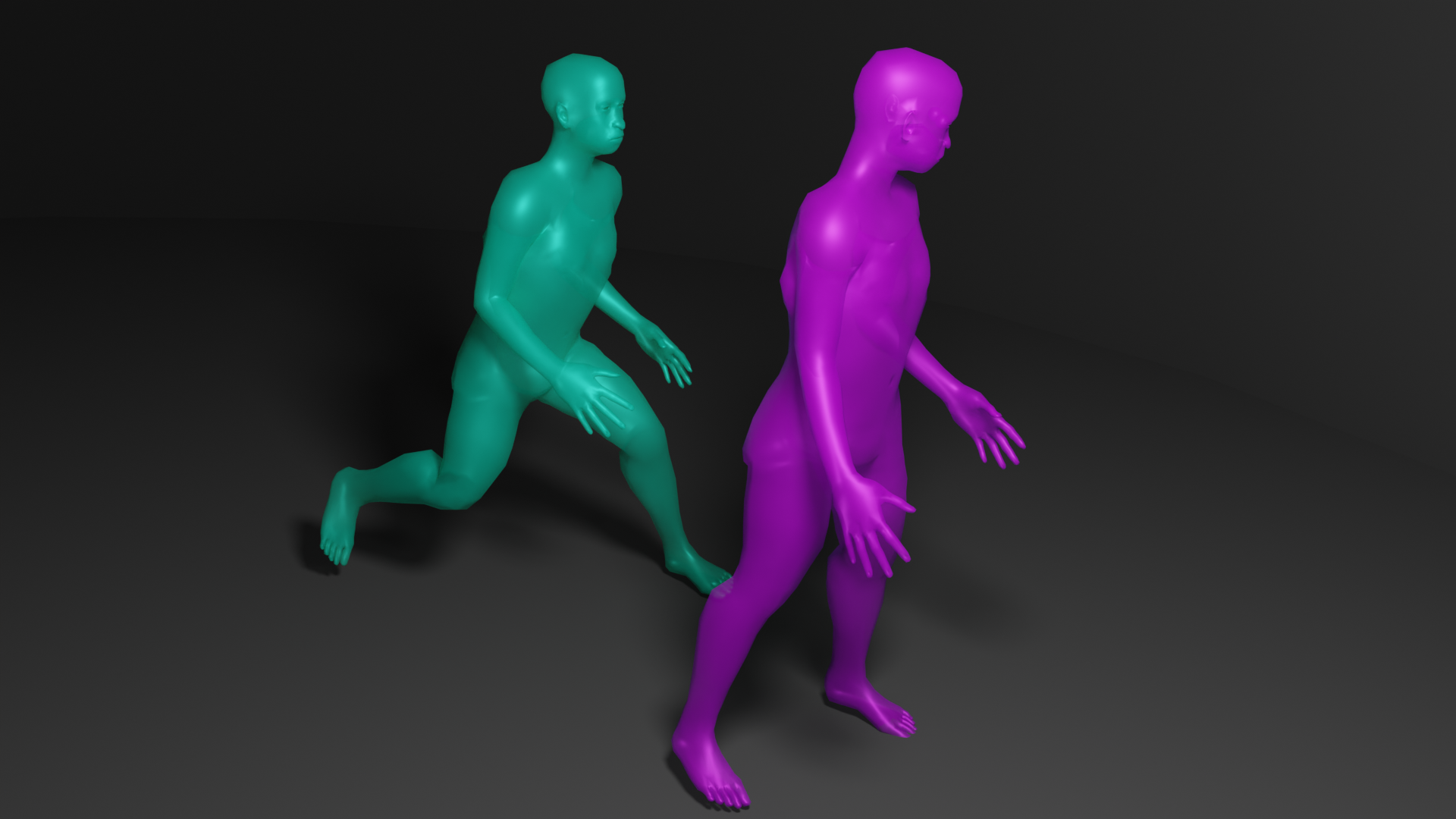}
  \includegraphics[width=\linewidth]{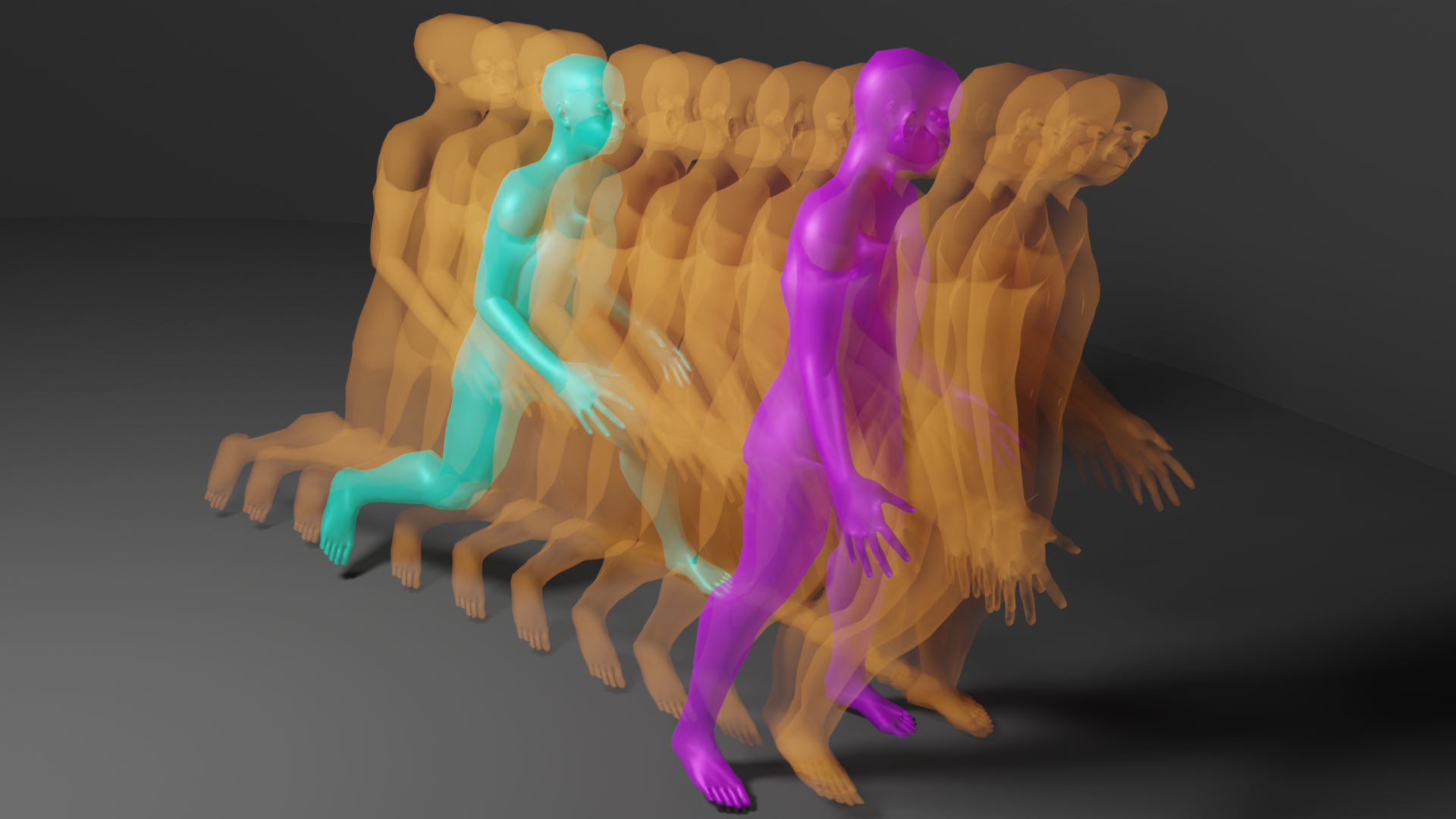}
  \caption{Qualitative Results: A sequence of a person jumping to the side while running.}
  \label{HMD:fig:Results_6}
\end{center}
\end{figure}

\begin{figure}[!p]
\begin{center}
  \includegraphics[width=\linewidth]{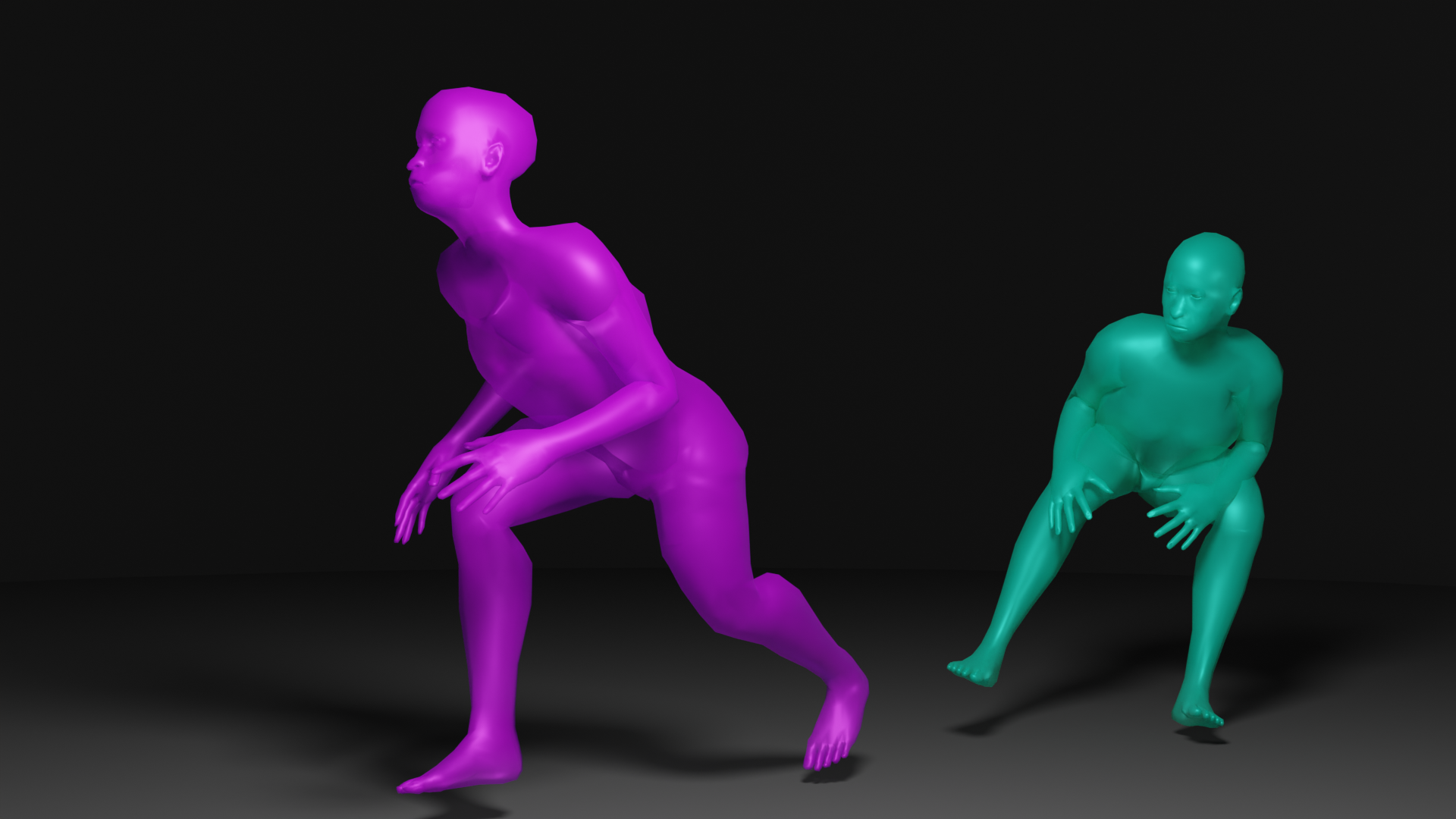}
  \includegraphics[width=\linewidth]{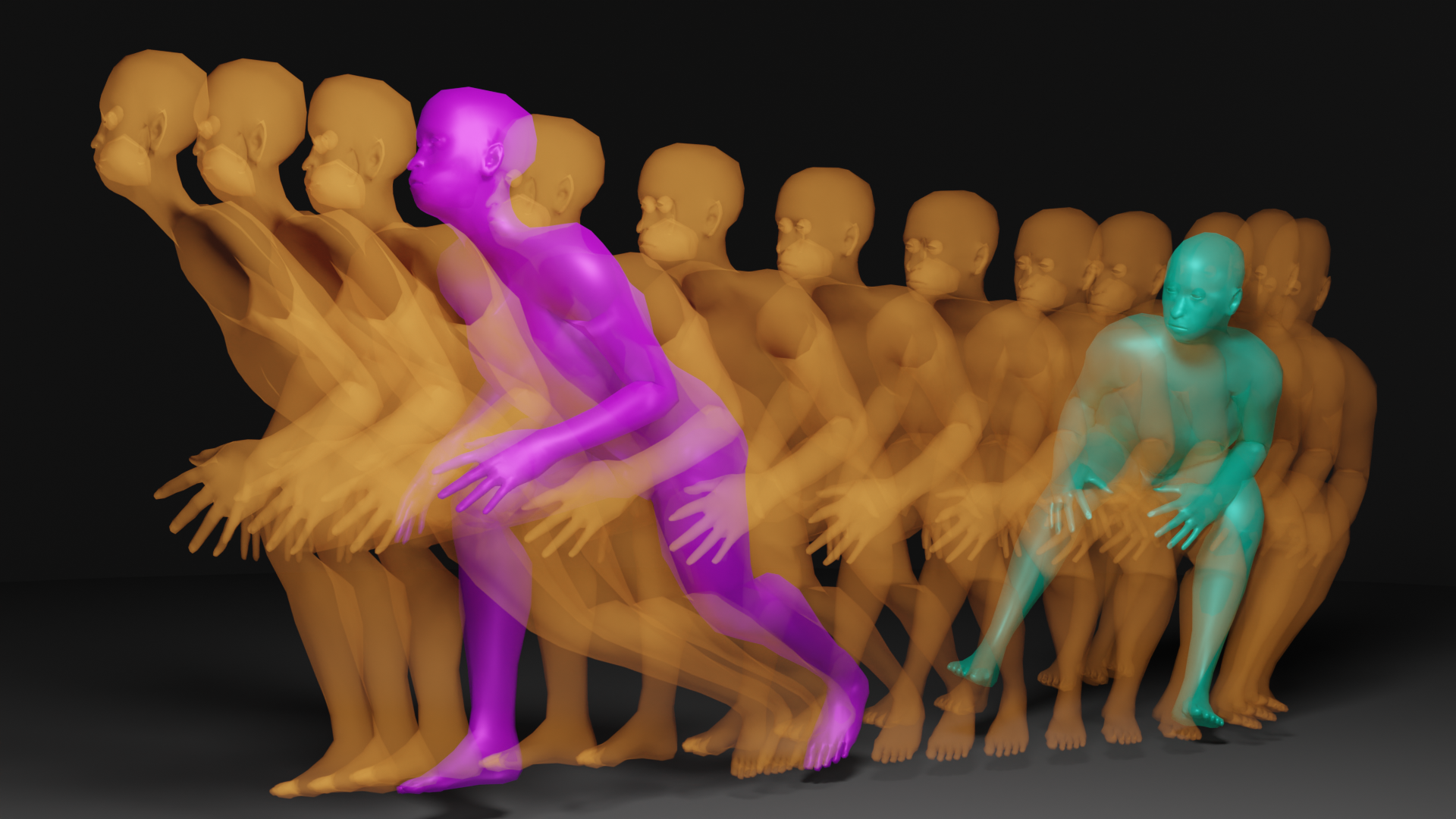}
  \caption{Qualitative Results: A sequence of a person performing complex movement involving ducking motion.}
  \label{HMD:fig:Results_7}
\end{center}
\end{figure}

The qualitative results for mesh blending obtained using our approach on the test set are shown in Fig. \ref{HMD:fig:Results_1}, \ref{HMD:fig:Results_2}, \ref{HMD:fig:Results_3}, \ref{HMD:fig:Results_4}, \ref{HMD:fig:Results_5}, \ref{HMD:fig:Results_6}, and \ref{HMD:fig:Results_7}. In each of these results, the first image shows the two input human meshes $\mathcal{G}_0$ and $\mathcal{G}_1$, and the second image shows the blending results for various values of $t$. 

In Fig. \ref{HMD:fig:Results_1}, the two input meshes have been sampled from a sequence where a person, while walking, bends forward and takes a right turn. The bending of the human and the movements of the legs while walking is visible in the results. In Fig. \ref{HMD:fig:Results_2}, a person moves their arms inwards while walking, initially with their arms outwards. This movement is successfully predicted by our method. In Fig. \ref{HMD:fig:Results_3}, the person is standing in one place and making complicated hand movements, which our method interpolates very realistically. Fig. \ref{HMD:fig:Results_4} takes the two inputs from a running person sequence. The hand motions are successfully predicted. However, since the right leg is in the same position in both the input, there is ambiguity about the number of steps the person takes, and our method fails to interpolate the movement of the legs. In Fig. \ref{HMD:fig:Results_5}, the input sequence has complex body motion, where the person is bending forward, with large movements in the hands and legs. Our method successfully interpolates this complex body movement. In Fig. \ref{HMD:fig:Results_6}, the sequence indicates a side jump while running. Our methods perform well in both the interpolations and the extrapolation tasks. Fig. \ref{HMD:fig:Results_7}, the input sequence contains a complex movement involving ducking motion. The results for both interpolation and extrapolation are realistic.

As can be seen, the movement of each part of the body is very realistic, for example, the movement of the hands, legs, and heads. Even in the case of complex input shapes, our network can create a smooth deformation across time, implying that the network has an implicit understanding of human motion to some extent.





\section{Conclusion}
\label{HMD:sec:conclusion}

We have proposed a self-supervised deep learning framework for solving the mesh blending problem in scenarios where the meshes are not in correspondence. To solve this problem, we have developed Red-Blue MPNN, a novel graph neural network that processes an augmented graph to estimate the correspondence. We have designed a novel conditional refinement scheme to find the exact correspondence when certain conditions are satisfied. We further develop a graph neural network that takes the aligned meshes and the time value as input and fuses this information to process further and generate the desired result. Using motion capture datasets and human mesh designing software, we create a large-scale synthetic dataset consisting of temporal sequences of human meshes in motion.
Our results demonstrate that the proposed approach generates realistic deformation of body parts given complex inputs. Although the movements of the local body parts are realistic, there is a gliding effect wherein the global translation of the meshes needs to be appropriately handled. We aim to address this limitation in our future work.

\clearpage
\newpage

\bibliographystyle{IEEEbib}
\bibliography{ref}

\end{document}